\documentclass[acmtog]{acmart}
\AtBeginDocument{%
  }

\copyrightyear{2026}
\acmYear{2026}
\setcopyright{cc}
\setcctype{by}
\acmConference[SA Conference Papers '26]{SIGGRAPH Asia 2026 Conference Papers}{December 01--04, 2026}{Kuala Lumpur, Malaysia}
\acmBooktitle{SIGGRAPH Asia 2026 Conference Papers (SA Conference Papers '26), December 01--04, 2026, Kuala Lumpur, Malaysia}
\acmDOI{10.1145/3829340.3842323}
\acmISBN{979-8-4007-2842-6/2026/12}

\acmSubmissionID{2148}

\usepackage{colortbl}
\newcommand{\cellbest}{\cellcolor{red!23}}
\newcommand{\cellsecond}{\cellcolor{orange!26}}

\definecolor{linkblue}{HTML}{1677B8}

\begin{document}

\title{WildAnySplat: Feed-Forward 3D Gaussian Splatting in the Wild}

\author{Richard Shaw}
\affiliation{%
  \institution{Huawei Noah's Ark Lab}
  \city{London}
  \country{UK}
}
\email{richard.shaw1@huawei.com}

\author{Arthur Moreau}
\affiliation{%
  \institution{Huawei Noah's Ark Lab}
  \city{London}
  \country{UK}}
\email{arthur.moreau3@huawei.com}

\author{Athanasios Papaioannou}
\affiliation{%
  \institution{Huawei Noah's Ark Lab}
  \city{London}
  \country{UK}
}
\email{athanasios.papaioannou@huawei.com}

\author{Thomas Tanay}
\affiliation{%
 \institution{Huawei Noah's Ark Lab}
  \city{London}
  \country{UK}
 }
\email{thomas.tanay@huawei.com}

\author{Zhang Zhensong}
\affiliation{%
  \institution{Huawei Noah's Ark Lab}
  \city{London}
  \country{UK}
  }
\email{zhangzhensong@huawei.com}

\author{Eduardo Pérez-Pellitero}
\affiliation{%
  \institution{Huawei Noah's Ark Lab}
  \city{London}
  \country{UK}
  }
\email{e.perez.pellitero@huawei.com}

\renewcommand{\shortauthors}{Shaw et al.}

\begin{abstract}
We present \textit{WildAnySplat}, a feed-forward 3D Gaussian Splatting framework that enables robust novel-view synthesis from in-the-wild multi-view images. Unlike controlled captures, real-world images often contain multi-view inconsistencies caused by transient occluders, illumination changes, and camera ISP, which degrade the quality of existing feed-forward reconstruction methods.

To address this challenge, we introduce a student--teacher framework that learns transient-robust and appearance-consistent 3D representations. A teacher network operates on clean images, while a student processes the corresponding in-the-wild inputs. The teacher provides clean supervision to guide the student toward a consistent canonical representation. Since paired clean and in-the-wild data are scarce, we use a hybrid data strategy, combining real paired captures, synthetic augmentations, and pseudo-clean labels generated from optimization-based reconstructions, bridging the domain gap between synthetic and real-world data.

We adopt a transformer architecture with dedicated DPT heads for camera pose, depth, and Gaussian prediction. We additionally predict per-view transient masks to reweight photometric supervision and suppress unreliable Gaussians. To account for cross-view appearance variation, we incorporate a bilateral grid module that models per-image color transformations and enables recovery of a canonical radiance representation.

Experiments on challenging in-the-wild benchmarks show that \textit{WildAnySplat} improves robustness to transient occlusions and appearance variation while retaining the fast inference of feed-forward Gaussian Splatting.
\end{abstract}


\begin{teaserfigure}
  \includegraphics[width=\textwidth]{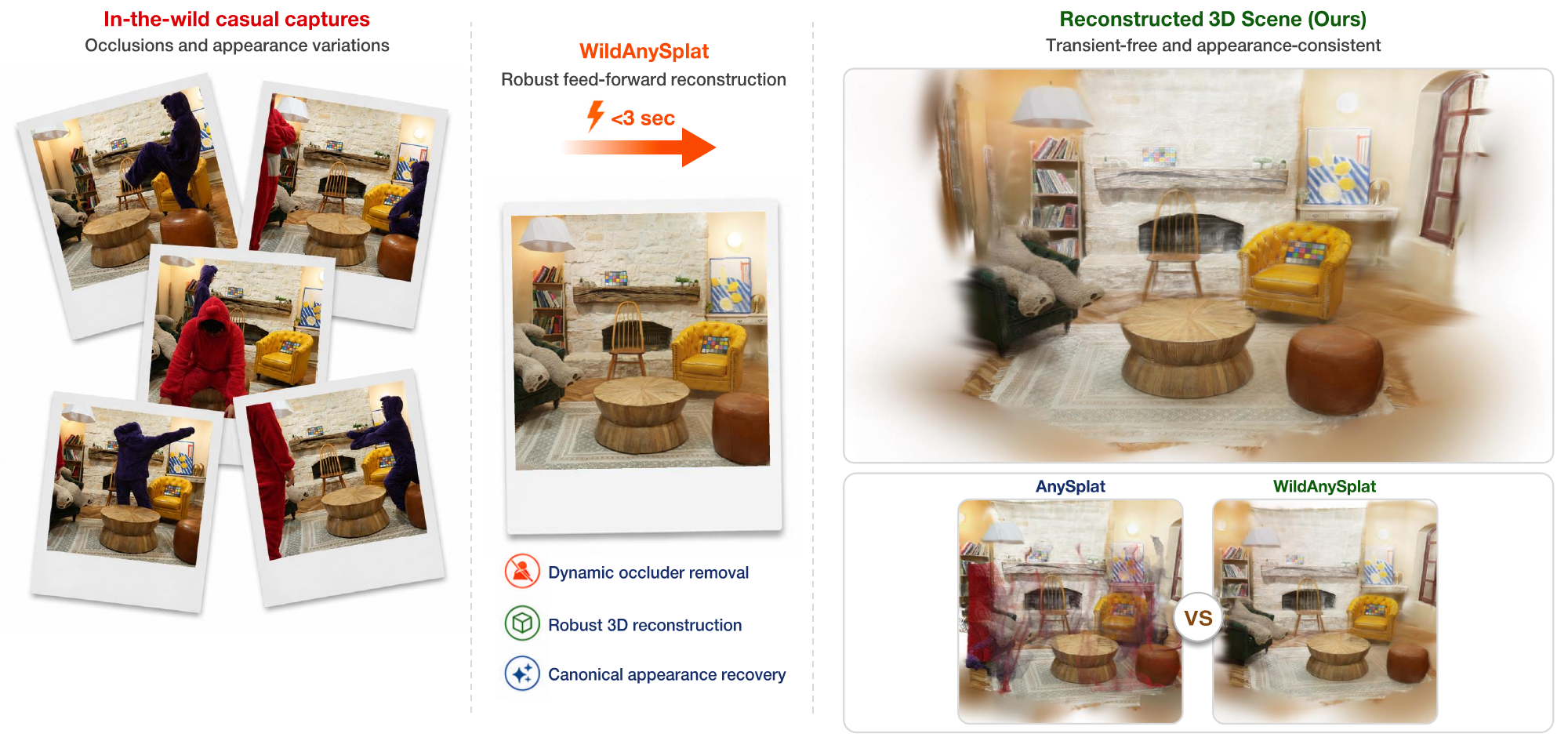}
\caption{
\textbf{WildAnySplat} is a feed-forward 3DGS framework for robust NVS from in-the-wild images with transient occluders, illumination changes, and camera-dependent appearance variation. We predict transient-aware Gaussians and per-view appearance transforms, enabling canonical scene representations with reduced ghosting and improved multi-view consistency. 
\textbf{Project page:}
    \href{https://wildanysplat.github.io/}
    {\textcolor{linkblue}{https://wildanysplat.github.io/}}
}
  \Description{WildAnySplat hero teaser figure.}
  \label{fig:teaser}
\end{teaserfigure}


\maketitle

\section{Introduction}
Feed-forward 3D reconstruction methods have recently made significant progress in enabling fast novel-view synthesis from multi-view images~\cite{wang2025vggt,Jiang2025AnySplatF3}. In particular, 3D Gaussian Splatting (3DGS)~\cite{3dgs} has emerged as a powerful representation due to its rendering efficiency and high visual fidelity, and various follow-ups have adapted foundational 3D models to decode Gaussian primitives~\cite{Jiang2025AnySplatF3, offthegrid}. However, existing feed-forward approaches largely assume consistent multi-view observations, limiting their applicability to controlled capture settings.

In contrast, real-world image collections, such as those found in internet photo datasets, exhibit substantial multi-view inconsistencies. These include transient occluders (e.g., pedestrians, vehicles), illumination changes across time, and camera-dependent color processing. Such effects violate the assumptions of multi-view consistency and lead to degraded geometry, color artifacts, and ghosting in reconstructed scenes.

While optimization-based methods such as NeRF in the Wild~\cite{martinbrualla2020nerfw} and WildGaussians~\cite{kulhanek2024wildgaussians} address these challenges through per-scene fitting and explicit robustness mechanisms, they incur high computational cost and are unsuitable for real-time or large-scale deployment. Feed-forward approaches, on the other hand, remain brittle under these conditions due to their inability to disentangle transient and persistent scene components.

In this work, we propose \textit{WildAnySplat}, a feed-forward framework that explicitly models and mitigates multi-view inconsistencies in in-the-wild images. Our method builds upon the AnySplat architecture and introduces three key components. First, we predict per-frame transient masks that identify unreliable regions, which are used both to reweight training losses and to guide voxelization fusion of Gaussians. Second, we model per-image color transformations via learned bilateral grids, enabling recovery of a consistent canonical appearance across views. Third, we use a student--teacher training framework with hybrid supervision, where a teacher network operating on clean observations provides stable geometric supervision to a student trained on in-the-wild inputs.

A central challenge in this setting is the absence of ground-truth supervision for clean, transient-free observations. To address this, we combine large-scale synthetic augmentation with pseudo supervision derived from optimization-based methods. Specifically, we generate synthetic transient artifacts and color perturbations during training, and further leverage outputs from methods such as WildGaussians to provide pseudo-clean images and uncertainty maps. This hybrid supervision strategy enables our model to generalize effectively to real-world in-the-wild data.

Our results demonstrate that WildAnySplat achieves improved robustness relative to the AnySplat backbone and competitive performance among current feed-forward in-the-wild methods, with the
largest quantitative gains occurring on dynamic RealX3D captures.


Our contributions are:
\begin{enumerate}
\item \textbf{Unified handling of in-the-wild inconsistencies in feed-forward 3DGS.} A student--teacher framework learns transient-robust and appearance-consistent canonical 3D representations from inconsistent multi-view images.

\item \textbf{Transient-aware Gaussian fusion via learned masks.} Per-view transient masks enable robust photometric supervision and suppress unreliable Gaussians during voxel fusion.

\item \textbf{Spatially-varying appearance modeling with bilateral grids.} Per-view bilateral grids model cross-view appearance variation, enabling recovery of a canonical appearance.

\item \textbf{Hybrid supervision with synthetic augmentation and pseudo labels.} Combining transient and ISP augmentation with pseudo labels from optimization methods addresses limited paired supervision and the synthetic-to-real domain gap.

\end{enumerate}

\subsection{Related Work}
\subsubsection{Optimization-based Novel View Synthesis in the Wild}

Opt-imization-based methods have been the dominant approach for handling in-the-wild image collections due to their flexibility in modeling inconsistencies. NeRF in the Wild~\cite{martinbrualla2020nerfw} introduced per-image appearance embeddings to account for illumination and color variation across images, with transient regions handled by uncertainty estimation. Subsequent methods further improved reconstruction quality, efficiency, and sparse inputs~\cite{mipnerf, zipnerf, refnerf, regnerf, nerfonthego, Lin2024VastGaussianV3}. 

Recent methods based on 3DGS explore similar approaches. Methods such as GS-W~\cite{zhang2024GS-W}, Wild-GS~\cite{Xu2024WildGSRN}, SWAG~\cite{Dahmani2024SWAGSI}, and WildGaussians~\cite{kulhanek2024wildgaussians} incorporate robust loss functions, visibility modeling, or uncertainty estimation to handle occlusions and inconsistent observations. SpotlessSplats~\cite{sabourgoli2024spotlesssplats} and DeGauss~\cite{wang2025degauss} explicitly address distractors by separating static and dynamic components or filtering transient artifacts during optimization. While these methods achieve strong performance, they rely on computationally expensive per-scene optimization and are therefore not suitable for real-time or large-scale applications.

\subsubsection{Feed-forward Novel View Synthesis}

Recent feed-forward approaches enable faster, generalizable reconstruction by learning mappings from input images to scene representations. Early works such as pixelNeRF~\cite{yu2021pixelnerf} and MVSNeRF~\cite{mvsnerf} learn generalizable radiance field predictors from sparse views. More recently, feed-forward approaches have been extended to 3DGS~\cite{offthegrid, pf3plat, zhang2025FLARE, Tian2025UniForwardU3, wang2025volsplat, pixelsplat, mvsplat, mvsgaussian}, enabling efficient inference with high rendering quality. Recent pose-free methods such as AnySplat~\cite{Jiang2025AnySplatF3} and DA3~\cite{depthanything3} demonstrate that transformer-based architectures can jointly predict camera poses, depths and Gaussian representations directly from multi-view inputs. Although somewhat robust to minor inconsistencies, these approaches still assume consistent multi-view observations and struggle with heavy occlusions and appearance variation.

\subsubsection{Feed-forward Novel View Synthesis in the Wild}

Several recent works extend feed-forward NVS methods to unconstrained settings. GenWildSplat~\cite{gupta2026genwildsplat} and Sparse-View 3DGS in the Wild~\cite{Li2025SparseGSWS3} address the challenge of generalization to real-world image collections with varying capture conditions, but focus on sparse views. WildSplatter~\cite{Fujimura_2026_WildSplatter} introduces appearance control mechanisms for feed-forward 3DGS, while WildRayZer~\cite{chen2026wildrayzer} explores self-supervised learning for large-scale dynamic environments. Wild3A~\cite{Li2025Wild3A} further demonstrates fast feed-forward reconstruction in dynamic scenes.

Despite this progress, existing methods typically address only subsets of the challenges present in in-the-wild data. Handling transient occluders and color inconsistencies are often treated independently, and supervision remains limited due to the lack of ground-truth clean observations. Notably, GenWildSplat uses a pretrained segmentation network~\cite{yolov8} to predict transient masks, rather than detecting occluders that are truly multi-view inconsistent. WildRayZer predicts motion masks and masks input tokens prior to entering the reconstruction network, rather than learning unified, robust 3D reconstruction at the representational level.

Our work differs from these approaches in three key aspects. First, our clean-teacher/in-the-wild-student training strategy maintains more robust reconstruction and pose estimation in the presence of inconsistent inputs. Second, we jointly model transient occluders and spatially-varying appearance from multi-view inputs via learned masks and bilateral grids, influencing both training and Gaussian fusion. Third, we introduce a hybrid supervision strategy that combines synthetic augmentation with pseudo ground-truth signals derived from optimization-based methods, bridging the gap between synthetic and real-world data.

\subsubsection{Appearance Modeling and Color Consistency}
Modeling appearance variation across views is critical for in-the-wild reconstruction. While optimization-based methods~\cite{martinbrualla2020nerfw, kulhanek2024wildgaussians, wang2024bilateral} typically learn per-image latent variables or affine color corrections jointly during reconstruction, recent work explores harmonizing images before or after reconstruction. PPISP~\cite{ppisp} uses a learned post-processing module for radiance field rendering that models common photometric variations, e.g. exposure, white-balance, CRF etc. Difix3D+~\cite{wu2025difix3d} enhances NVS renderings through single-step diffusion. \cite{alzayer2025generativemvr} employ a diffusion-based relighting model to match images with illumination variations to a given reference image prior to NeRF reconstruction. Similarly, CHROMA~\cite{shin2023chroma} proposes feed-forward bilateral grid prediction for consistent multi-view appearance harmonization prior to scene-specific 3DGS. We build on this idea by integrating bilateral grid prediction into a fully feed-forward 3DGS framework, allowing us to recover canonical Gaussian colors while compensating for per-image appearance variation.



\section{Method}

\subsection{Overview}

\begin{figure*}[t]
  \centering
  \includegraphics[width=\textwidth]{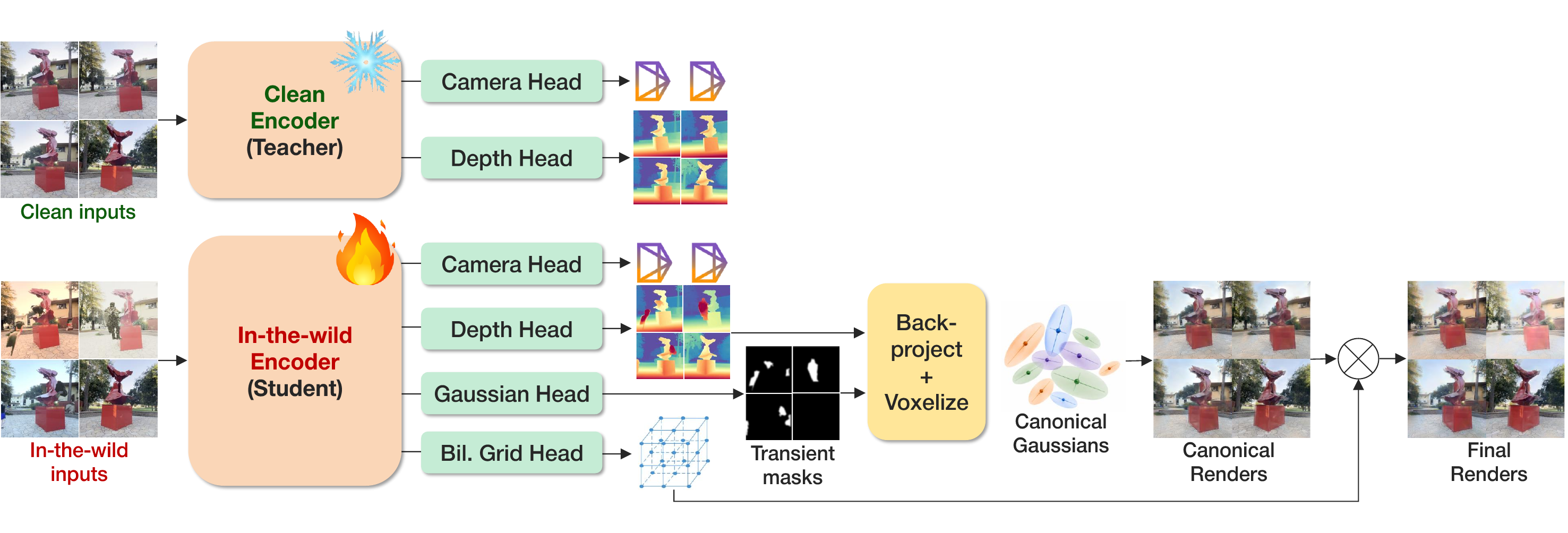}
    \caption{
    \textbf{Method overview.} WildAnySplat trains a student network to reconstruct feed-forward 3D Gaussians from in-the-wild images containing transient occluders and appearance variation. A frozen clean-image teacher provides camera and depth supervision, while the student predicts Gaussians, transient masks, and per-view bilateral grids. The transient masks downweight unreliable regions during photometric training and voxel fusion, while bilateral grids map canonical Gaussian renderings into each input image's appearance domain, enabling robust geometry and consistent canonical colors.
    }
  \label{fig:method}
\end{figure*}

Given a set of in-the-wild multi-view images $\{I_i\}$ exhibiting significant cross-view inconsistencies, our goal is to predict a 3D Gaussian representation that enables high-quality novel-view synthesis.

To address this challenge, we propose \textit{WildAnySplat}, a feed-forward 3DGS framework trained using a student--teacher paradigm. The student network operates on in-the-wild inputs, while a frozen teacher network processes corresponding clean observations obtained from synthetic augmentation, optimization-based pseudo-clean reconstructions, and a small set of real-world paired captures. The teacher provides stable geometric supervision in the form of camera poses and depth predictions, guiding the student toward more robust geometry under real-world inconsistencies.

Our framework consists of three key components: 1) feed-forward Gaussian prediction, 2) transient-aware masking and Gaussian voxel fusion, and 3) per-view appearance modeling via learned bilateral grids, all trained under the student--teacher framework. An overview of our framework is shown in Fig.~\ref{fig:method}.

\subsection{Student--Teacher Framework}

Let $\{I_i^{w}\}$ denote in-the-wild input images and $\{I_i^{c}\}$ their corresponding clean counterparts. Clean images are obtained via a combination of real-world paired captures, synthetic augmentations, and optimization-based reconstruction (e.g., WildGaussians~\cite{kulhanek2024wildgaussians}).

We use two networks with a shared VGGT~\cite{wang2025vggt} backbone architecture:

\begin{itemize}
    \item a \textbf{teacher} network $f_{\theta^t}$ operates on clean images $\{I_i^{c}\}$,
    \item a \textbf{student} network $f_{\theta^s}$ operates on in-the-wild images $\{I_i^{w}\}$.
\end{itemize}

Both networks are equipped with DPT heads~\cite{dpthead} to predict camera poses $\mathcal{C}$ and depth maps $\mathcal{D}$. The student network has additional DPT heads to predict per-pixel Gaussians $\mathcal{G}$, transient masks $\mathcal{T}$, and per-view bilateral grids $\mathcal{B}$:
\begin{equation}
\mathcal{C}^t, \mathcal{D}^t = f_{\theta^t}(\{I_i^{c}\}), \quad
\mathcal{C}^s, \mathcal{D}^s, \mathcal{G}^s, \mathcal{T}^s, \mathcal{B}^s = f_{\theta^s}(\{I_i^{w}\}).
\end{equation}

As the teacher operates only on clean images, it provides stable camera pose and depth predictions that are used to supervise the student via geometric distillation losses.

\paragraph{Student Guidance.}
The student is supervised using a transient-masked photometric loss on the in-the-wild inputs together with a full photometric loss on the corresponding clean images:
\begin{equation}
\mathcal{L}_{\text{photo}} = \lambda_c \mathcal{L}_{\text{photo}^c} + \lambda_w \mathcal{L}_{\text{photo}^w}
\end{equation}
In addition to photometric losses, we enforce consistency with the camera pose and depth outputs from the clean teacher:
\begin{equation}
\mathcal{L}_{\text{distill}} = \mathcal{L}_{\mathcal{C}} + \mathcal{L}_{\mathcal{D}}
\end{equation}
The photometric and distillation losses form the primary supervision terms for student training.

\subsubsection{Feed-forward Gaussian Prediction}

Following prior work~\cite{Jiang2025AnySplatF3}, the student network uses a dedicated DPT head to predict a set of pixel-aligned 3D Gaussians. Each Gaussian $g$ is parameterized by mean $\boldsymbol{\mu} \in \mathbb{R}^3$, covariance $\boldsymbol{\Sigma} \in \mathbb{R}^{3 \times 3}$, opacity $\alpha$, and color $\mathbf{c}$. Per-view features are extracted and fused via cross-attention, producing a latent representation that is decoded into Gaussians. The per-view pixel-aligned Gaussians are subsequently fused via voxelization and rendered using standard alpha compositing~\cite{ye2025gsplat}.

\subsubsection{Transient Mask Prediction}
\label{sec:masks}

To handle transient occluders and inconsistent observations, the student predicts a static probability map for each input image $I_i^w$. Low probabilities correspond to transient or unreliable regions, while high probabilities correspond to persistent scene content. Specifically, we extend the Gaussian DPT head with an additional output channel corresponding to the probability that each lifted point or Gaussian belongs to the persistent static scene. We denote this predicted static probability by
\begin{equation}
    M_i(p) \in [0,1],
\end{equation}
where $p$ indexes an input pixel and its corresponding lifted 3D point. Intuitively, $M_i(p)$ should be close to one for pixels observing stable scene content and close to zero for transient occluders such as pedestrians, vehicles, or dynamic scene elements.

When ground-truth or pseudo transient labels are available, we supervise the predicted masks using a positive-weighted binary cross-entropy loss:
\begin{equation}
    \mathcal{L}_{\text{mask}}
    =
    \operatorname{BCE}(M_i, Y_i; \beta),
\end{equation}
where $Y_i$ denotes the target static mask and positive weighting compensates for the strong class imbalance between static and transient pixels.

The predicted static probabilities are subsequently used in both photometric supervision and voxel fusion. During training, transient regions are downweighted in the photometric loss, encouraging the network to focus on multi-view consistent observations. During voxel fusion, the predicted static probabilities modulate the contribution of per-view Gaussians, suppressing transient observations when constructing the fused scene representation.

\paragraph{Masked Photometric Loss.}
The predicted static probabilities are used to reweight the photometric reconstruction loss:
\begin{equation}
\mathcal{L}_{\text{photo}^w} =
\sum_i \sum_p
M_i(p)
\left\|
\hat{I}_i^s(p) - I_i^{w}(p)
\right\|_1,
\end{equation}
where $\hat{I}_i^s$ denotes the rendering of the student-predicted Gaussians from the camera of image $I_i^w$. This encourages the model to focus photometric supervision on static, multi-view consistent regions while downweighting transient occluders.

\paragraph{Mask-weighted Voxel Fusion.}
In addition to photometric loss reweighting, we incorporate the predicted static probabilities into the voxel fusion stage used to merge per-view Gaussian predictions into a unified scene representation.

For each lifted point $p$ from view $i$, let $\mathbf{x}_{i,p}$ denote its 3D position, $\mathbf{f}_{i,p}$ its feature representation, $\mathrm{conf}_{i,p}$ its confidence score, and $M_i(p)$ its predicted static probability. We discretize points into voxels and aggregate per-view Gaussian predictions using transient-aware confidence weighting:
\begin{equation}
    w_{i,p}
    \propto
    \exp\left(\frac{\mathrm{conf}_{i,p}}{\tau}\right)
    M_i(p)
    R_i(p),
\end{equation}
where $\tau$ is a temperature parameter and $R_i(p)\in[0,1]$ is an optional reliability term derived from depth validity or confidence filtering.

Let $\mathcal{P}_v$ denote the set of points assigned to voxel $v$. Voxel positions and features are then computed via weighted averaging:
\begin{equation}
    \mathbf{x}_v
    =
    \sum_{(i,p)\in\mathcal{P}_v}
    w_{i,p}\mathbf{x}_{i,p},
    \qquad
    \mathbf{f}_v
    =
    \sum_{(i,p)\in\mathcal{P}_v}
    w_{i,p}\mathbf{f}_{i,p}.
\end{equation}
By incorporating transient-aware weighting during voxel fusion, the model suppresses inconsistent observations when merging per-view Gaussian predictions, reducing ghosting artifacts and improving the stability of the final fused scene representation.

\subsubsection{Appearance Modeling with Bilateral Grids}

To model cross-view appearance variation, the student network employs an additional DPT head to predict a 3D bilateral grid $B_i$ for each input image $I_i^w$. Rather than directly predicting image-specific Gaussian colors, the student model maintains a \emph{canonical} Gaussian appearance representation and learns view-dependent color transformations that map canonical renderings into the appearance space of each input image. Specifically, we first render the canonical Gaussians to obtain a canonical image $\hat{I}_i^{canon}$, and then apply the predicted bilateral grid:
\begin{equation}
    \hat{I}_i^{w}
    =
    B_i(\hat{I}_i^{canon}).
\end{equation}

Following prior work on bilateral-grid appearance modeling~\cite{shin2023chroma}, each grid represents a spatially-varying affine color transformation:
\begin{equation}
\tilde{\mathbf{c}} =
\mathbf{A}_i(x,y,l)\,\mathbf{c}
+
\mathbf{b}_i(x,y,l),
\end{equation}
where $(x,y)$ are low-resolution spatial coordinates and $l$ is a luminance dimension. The predicted bilateral grids are subsequently \textit{sliced}~\cite{gharbi2017deep} using the canonical image $\hat{I}_i^{canon}$ to obtain per-pixel affine transforms.

To prevent degenerate solutions where the canonical color space becomes unnatural and the appearance modeling is absorbed entirely by the bilateral-grid transforms, we predict \emph{residual} affine transformations relative to the identity transform:
\begin{equation}
\mathbf{A}_i = \mathbf{I} + \Delta \mathbf{A}_i,
\qquad
\mathbf{b}_i = \Delta \mathbf{b}_i.
\end{equation}

This residual parameterization encourages the bilateral grids to model only image-specific appearance deviations while preserving stable canonical Gaussian colors. We regularize the predicted bilateral grids using total variation and residual identity penalties:
\begin{equation}
\mathcal{L}_{\text{bg}}
=
\lambda_{\text{tv}}
\sum_i \operatorname{TV}(B_i)
+
\lambda_{\text{id}}
\sum_i
\left\|
\Delta B_i
\right\|_1.
\end{equation}

By decoupling canonical scene appearance from image-specific color transforms, the bilateral-grid formulation enables robust handling of illumination changes, exposure variation, and camera ISP while preserving a consistent underlying Gaussian representation.

\subsubsection{Training Objective}

The student network is trained using the following overall loss function:
\begin{equation}
\mathcal{L} = 
\lambda_{\text{photo}} \mathcal{L}_{\text{photo}} +
\lambda_{\text{distill}} \mathcal{L}_{\text{distill}} +
\lambda_{\text{mask}} \mathcal{L}_{\text{mask}} +
\lambda_{\text{bg}} \mathcal{L}_{\text{bg}},
\end{equation}
where the teacher uses pretrained VGGT weights and remains frozen during training.

\subsection{Training Data}

We train on a diverse mixture of synthetic and real-world datasets following recent feed-forward methods~\cite{wang2025vggt, Jiang2025AnySplatF3}. We use scenes from 13 public datasets:
DL3DV~\cite{dl3dv}, 
Hypersim~\cite{hypersim},
ARKitScenes~\cite{arkitscenes},
RealEstate10K~\cite{re10k},
BlendedMVS~\cite{blendedmvs},
ScanNet++~\cite{scannetpp},
Co3D~\cite{co3d},
Unreal4K~\cite{unreal4k},
KITTI-360~\cite{kitti360},
DTU~\cite{dtu},
ETH3D~\cite{eth3d},
WildRGBD~\cite{wildrgbd}, and Dynamic-RE10K~\cite{chen2026wildrayzer}, as well as hold-out scenes from RealX3D~\cite{realx3d} and Phototourism~\cite{phototourism}. These span indoor and outdoor scenes, object-centric and large-scale captures, as well as a range of geometric and appearance variations. This diversity is important for robust in-the-wild reconstruction, as it exposes the model to varying scene layouts, illumination conditions, and camera characteristics.

A major challenge in training robust in-the-wild reconstruction models is the lack of paired supervision consisting of corrupted input images and corresponding clean observations. To address this, we generate clean and in-the-wild pairs using synthetic augmentation (transient object and camera ISP variation) and pseudo-clean supervision from optimization-based reconstruction as follows.

\subsubsection{Synthetic Transient \& ISP Augmentation}

To simulate transient occluders commonly observed in in-the-wild images, we augment clean training images by randomly inserting 2D objects sampled from the COCO dataset~\cite{coco}. Specifically, we randomly select segmented foreground objects and alpha-composite them into the training views with randomized scale, position, and appearance perturbations. For each augmented image, we additionally generate ground-truth transient masks corresponding to the inserted objects, which are used to supervise the transient mask prediction (Sec.~\ref{sec:masks}).

In practice, we found that naively pasting segmented objects into images produced unrealistic artifacts that were too easy for the network to detect. To improve realism and better approximate real-world transient occluders, we additionally apply several appearance adaptation operations to the pasted objects, including boundary blurring, noise injection, and affine color matching to the local background region. These augmentations encourage the model to learn semantic transient reasoning rather than relying on low-level copy-paste cues.

To further improve robustness to appearance inconsistencies, we apply randomized ISP-style color augmentations, including exposure shifts, white balance changes, saturation variation, contrast perturbations, vignetting, and local color transforms. Together, these augmentations simulate the lighting and camera-processing variations commonly encountered in casually captured imagery. Examples of our synthetic paired training data are shown in Fig.~\ref{fig:aug_examples}.

\subsubsection{Pseudo Clean Image Supervision}

\begin{figure}
    \centering
    \includegraphics[width=1.0\linewidth]{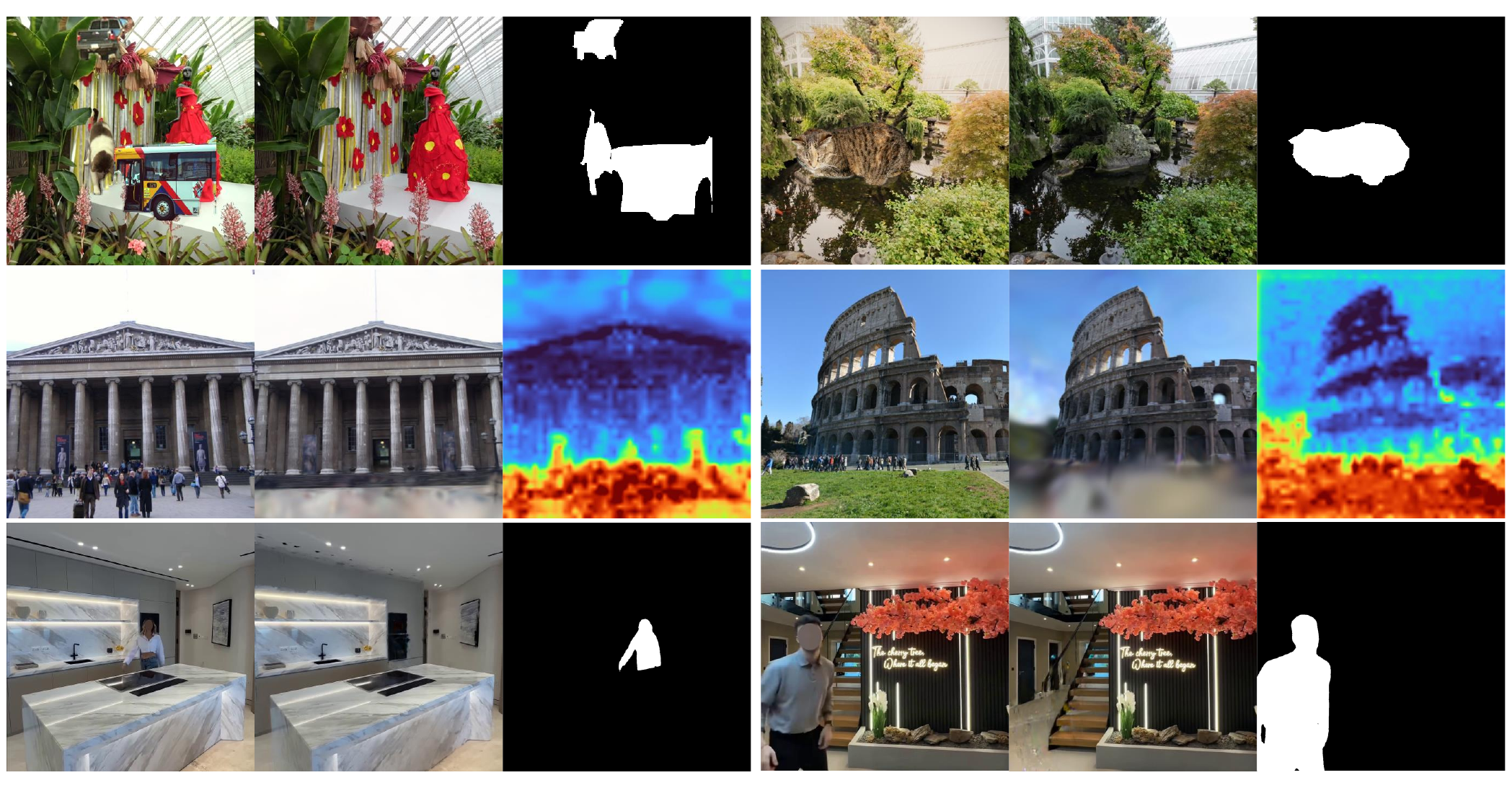}
    \caption{
    \textbf{Training examples.} We use a hybrid mix of synthetic augmentation, pseudo-paired data, and real-world pairs.
    Left to right: in-the-wild input images, GT clean or pseudo-clean images, and transient/uncertainty masks.
    }
    \label{fig:aug_examples}
\end{figure}

 To help close the domain gap between real and synthetic data, we construct an additional dataset of paired in-the-wild and pseudo-clean images using hold-out scenes from the Phototourism~\cite{phototourism} and Dynamic-RE10K~\cite{chen2026wildrayzer} datasets. Specifically, we employ WildGaussians~\cite{kulhanek2024wildgaussians} and render transient-suppressed pseudo-clean images from the resulting reconstructions. For each training image $I_i^w$, we obtain 1) a pseudo-clean target image $I_i^c$ rendered from the optimized WildGaussians reconstruction, and 2) uncertainty maps indicating unreliable or poorly reconstructed regions, e.g. transient regions. These pseudo-clean renderings serve as teacher supervision within our student--teacher framework, enabling the student network to learn transient-invariant canonical scene representations while remaining exposed to realistic in-the-wild degradations. Importantly, the scenes used for pseudo-clean supervision are disjoint from the evaluation scenes used in our experiments. Examples of our pseudo-clean paired data are shown in Fig.~\ref{fig:aug_examples}.

\subsubsection{Implementation Details}

We train on four GPUs with
140 GB of VRAM for 35k iterations, using AdamW optimizer and cosine learning rate scheduler, with a peak learning rate of 2e-5 and a warmup of 2k iterations. As in-the-wild data and heavy augmentation can destabilize training, we accumulate gradients over every 4 batches. Training datasets are sampled at random every iteration and we use a dynamic batch-size sampler that randomly samples between 2 and 16 frames, with a maximum input resolution of $448 \times 448$ px. For the training objective, we set $\lambda_{photo} = 1.0$, $\lambda_{distill} = 0.1$, and $\lambda_{bg} = 0.0001$. We predict bilateral grids at a resolution of $16 \times 16 \times 8$. The teacher network uses pretrained VGGT weights and remains frozen throughout training. At inference time, only the student network is used, enabling efficient feed-forward reconstruction. Additional implementation details, component ablations, and measurements of
inference runtime, peak GPU memory, and preprocessing cost are
provided in the supplementary material.

\begin{table*}[h]
\centering

\caption{Novel-view synthesis results on four benchmarks using 3, 6, 9, and 12 input views. WildAnySplat performs strongly compared to prior feed-forward methods, particularly on challenging scenes with transient occluders and appearance variation.}
\label{tab:nvs_results}

\small
\setlength{\tabcolsep}{2.0pt}
\resizebox{\textwidth}{!}{
\begin{tabular}{l|ccc|ccc|ccc|ccc||l|ccc|ccc|ccc|ccc}
\toprule
& \multicolumn{12}{c||}{\textbf{RealX3D}} 
& & \multicolumn{12}{c}{\textbf{Phototourism}} \\
& \multicolumn{3}{c|}{3 Views} & \multicolumn{3}{c|}{6 Views} & \multicolumn{3}{c|}{9 Views} & \multicolumn{3}{c||}{12 Views} 
& & \multicolumn{3}{c|}{3 Views} & \multicolumn{3}{c|}{6 Views} & \multicolumn{3}{c|}{9 Views} & \multicolumn{3}{c}{12 Views} \\
Method 
& PSNR$\uparrow$ & SSIM$\uparrow$ & LPIPS$\downarrow$ & PSNR$\uparrow$ & SSIM$\uparrow$ & LPIPS$\downarrow$ & PSNR$\uparrow$ & SSIM$\uparrow$ & LPIPS$\downarrow$ & PSNR$\uparrow$ & SSIM$\uparrow$ & LPIPS$\downarrow$ 
& Method 
& PSNR$\uparrow$ & SSIM$\uparrow$ & LPIPS$\downarrow$ & PSNR$\uparrow$ & SSIM$\uparrow$ & LPIPS$\downarrow$ & PSNR$\uparrow$ & SSIM$\uparrow$ & LPIPS$\downarrow$ & PSNR$\uparrow$ & SSIM$\uparrow$ & LPIPS$\downarrow$ \\
\midrule

AnySplat 
& 13.44 & 0.362 & 0.494 & 15.71 & 0.450 & 0.418 & 16.40 & 0.486 & 0.396 & 16.84 & 0.501 & 0.378 
& AnySplat 
& 11.28 & 0.358 & 0.538 & 13.93 & 0.414 & 0.469 & 13.80 & 0.413 & 0.484 & 14.17 & 0.396 & 0.492 \\

Off The Grid 
& 14.34 & \cellsecond 0.403 & 0.501 & 15.00 & 0.467 & 0.463 & 14.77 & 0.471 & 0.467 & 15.15 & 0.467 & 0.468 
& Off The Grid 
& 11.73 & 0.330 & 0.623 & 12.58 & 0.338 & 0.582 & 12.37 & 0.337 & 0.596 & 12.73 & 0.333 & 0.596 \\

WildSplatter 
& \cellsecond 14.97 & \cellbest 0.493 & \cellbest 0.435 & \cellsecond 17.69 & \cellbest 0.572 & \cellsecond 0.370 & \cellsecond 17.96 & \cellbest 0.579 & \cellsecond 0.364 & \cellsecond 17.59 & \cellbest 0.576 & \cellsecond 0.372 
& WildSplatter 
& 12.38 & \cellsecond 0.399 & \cellbest 0.496 & \cellbest 15.12 & \cellbest 0.490 & \cellbest 0.424 & \cellbest 16.17 & \cellbest 0.497 & \cellbest 0.439 & \cellbest 15.75 & \cellbest 0.491 & \cellbest 0.448 \\

GenWildSplat 
& 13.66 & 0.385 & 0.526 & 14.14 & 0.416 & 0.513 & 13.86 & 0.420 & 0.524 & 13.62 & 0.405 & 0.546 
& GenWildSplat 
& \cellbest 12.57 & \cellbest 0.405 & 0.562 & 14.69 & 0.429 & 0.526 & \cellsecond 14.71 & \cellsecond 0.430 & 0.534 & 13.97 & 0.406 & 0.550 \\

\textbf{WildAnySplat} 
& \cellbest 15.37 & 0.402 & \cellsecond 0.462 & \cellbest 18.02 & \cellsecond 0.507 & \cellbest 0.333 & \cellbest 19.83 & \cellsecond 0.576 & \cellbest 0.270 & \cellbest 19.56 & \cellsecond 0.556 & \cellbest 0.275 
& \textbf{WildAnySplat} 
& \cellbest 12.57 & 0.394 & \cellsecond 0.529 & \cellsecond 14.99 & \cellsecond 0.468 & \cellsecond 0.440 & 14.68 & 0.425 & \cellsecond 0.475 & \cellsecond 15.32 & \cellsecond 0.476 & \cellsecond 0.459 \\

\midrule
\midrule

& \multicolumn{12}{c||}{\textbf{NeRF On-the-Go}} 
& & \multicolumn{12}{c}{\textbf{RobustNeRF}} \\
\midrule

AnySplat 
& 10.48 & 0.245 & \cellsecond 0.624 & 12.36 & 0.260 & 0.582 & 13.14 & 0.290 & \cellsecond 0.544 & 14.14 & 0.307 & \cellsecond 0.529 
& AnySplat 
& 10.05 & 0.344 & 0.668 & 12.00 & 0.403 & 0.621 & 14.28 & 0.461 & \cellbest 0.554 & 14.00 & 0.459 & \cellsecond 0.574 \\

Off The Grid 
& \cellbest 13.64 & \cellbest 0.322 & 0.629 & \cellbest 13.91 & \cellsecond 0.325 & 0.584 & \cellbest 14.54 & \cellbest 0.347 & 0.559 & \cellsecond 14.76 & \cellsecond 0.355 & 0.543 
& Off The Grid 
& \cellbest 13.64 & \cellbest 0.484 & \cellsecond 0.641 & \cellbest 13.98 & \cellbest 0.499 & 0.627 & 14.44 & \cellbest 0.513 & 0.589 & 14.44 & \cellbest 0.518 & 0.599 \\

WildSplatter 
& \cellsecond 11.46 & 0.284 & \cellbest 0.600 & \cellsecond 13.45 & \cellbest 0.338 & \cellbest 0.552 & 13.74 & \cellsecond 0.337 & 0.565 & 14.71 & \cellbest 0.360 & 0.552 
& WildSplatter 
& \cellsecond 13.18 & 0.386 & \cellbest 0.620 & \cellsecond 13.97 & 0.448 & \cellbest 0.599 & \cellbest 14.83 & 0.481 & 0.592 & \cellsecond 14.53 & 0.477 & 0.610 \\

GenWildSplat 
& 10.91 & 0.271 & 0.640 & 11.50 & 0.266 & 0.637 & 12.12 & 0.284 & 0.620 & 12.33 & 0.293 & 0.616 
& GenWildSplat 
& 9.25 & 0.355 & 0.695 & 10.93 & 0.394 & 0.663 & 12.28 & 0.417 & 0.640 & 12.33 & 0.433 & 0.640 \\

\textbf{WildAnySplat} 
& \cellsecond 11.46 & \cellsecond 0.285 & 0.625 & 13.59 & 0.312 & \cellsecond 0.561 & \cellsecond 14.19 & 0.320 & \cellbest 0.528 & \cellbest 15.24 & 0.343 & \cellbest 0.492 
& \textbf{WildAnySplat} 
& 11.97 & \cellsecond 0.444 & 0.650 & 13.10 & \cellsecond 0.468 & \cellsecond 0.603 & \cellsecond 14.77 & \cellsecond 0.505 & \cellsecond 0.563 & \cellbest 14.78 & \cellsecond 0.509 & \cellbest 0.559 \\

\bottomrule
\end{tabular}
}

\end{table*}

\section{Experiments}

\paragraph{Datasets.} We evaluate WildAnySplat on four real-world in-the-wild NVS benchmarks:

\begin{itemize}
    \item \textbf{RealX3D}~\cite{realx3d}: High-quality real-world paired multi-view captures with dynamic occluders.
    \item \textbf{Phototourism}~\cite{phototourism}: Internet photo collections with severe appearance variation and transient occluders.
    \item \textbf{NeRF On-the-Go}~\cite{nerfonthego}: Casually captured scenes containing dynamic objects and lighting changes.
    \item \textbf{RobustNeRF}~\cite{robustnerf}: Challenging real-world scenes with strong multi-view inconsistencies.
\end{itemize}

Following prior work, we evaluate NVS using $3$, $6$, $9$ and $12$ input views. Performance is measured using PSNR, SSIM, and LPIPS.

\paragraph{Baselines.} We compare to recent feed-forward 3DGS methods:
\begin{itemize}
\item \textbf{AnySplat}~\cite{Jiang2025AnySplatF3}: The backbone on which our method is built. Extends VGGT~\cite{wang2025vggt} for feed-forward 3DGS and aggregates pixel-aligned Gaussians in a voxel grid.
\item \textbf{Off The Grid}~\cite{offthegrid}: Replaces pixel-aligned Gaussians with adaptive sub-pixel placement. Usually outperforms AnySplat, but uses colors from source images for Gaussian colors, making it weak under appearance variation.
\item \textbf{WildSplatter}~\cite{Fujimura_2026_WildSplatter}: Jointly predicts Gaussians and embeddings for unconstrained images, enabling flexible modulation of Gaussian colors for appearance control.
\item \textbf{GenWildSplat}~\cite{gupta2026genwildsplat}: Predicts depth, cameras, and Gaussians in a canonical space from sparse, unposed images. Incorporates learned appearance and semantic segmentation to handle illumination changes and transients.
\end{itemize}

\subsection{Novel View Synthesis Results}

\begin{figure*}[t]
    \centering
    \includegraphics[width=0.75\textwidth]{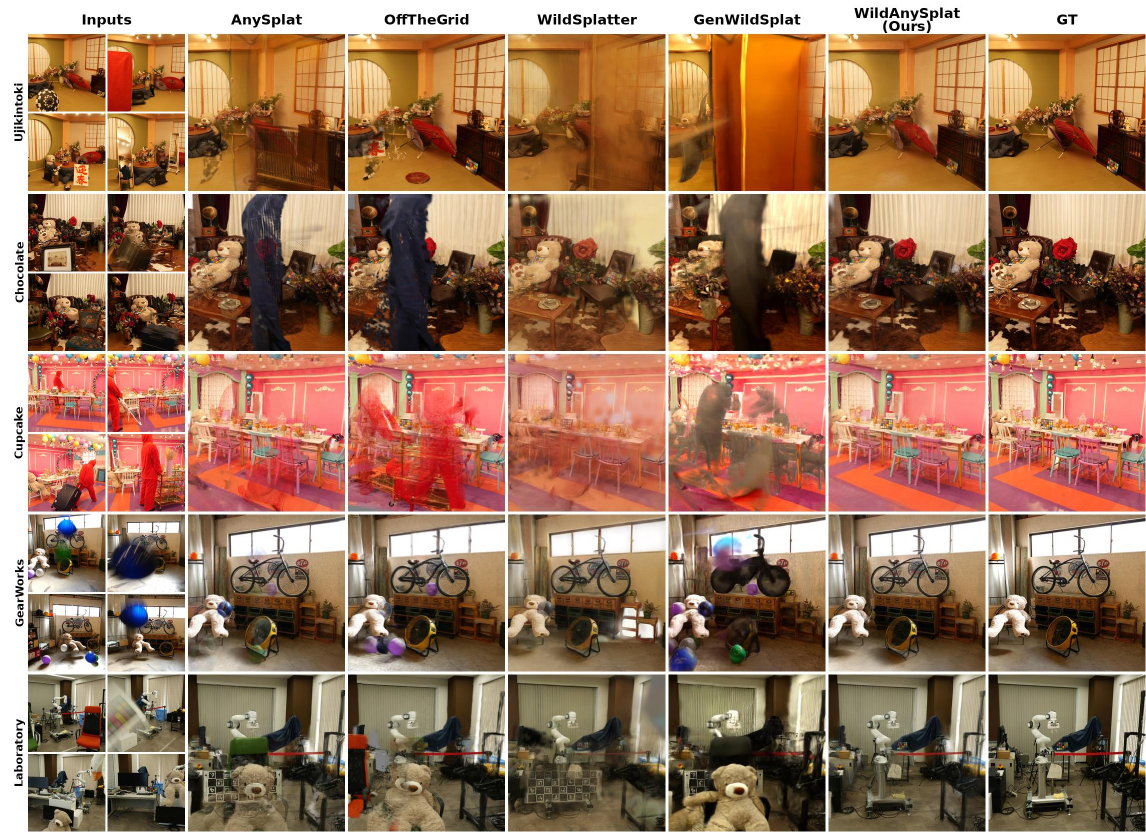}
\caption{
Qualitative comparison on in-the-wild scenes from the RealX3D~\cite{realx3d} dataset. WildAnySplat produces novel-view renderings with fewer transient artifacts and more consistent geometry than prior methods.}
\label{fig:qualitative_grid_realx3d}
\end{figure*}

Quantitative results are presented in Table~\ref{tab:nvs_results}.
WildAnySplat achieves competitive performance across the evaluated datasets and
input-view settings, with its largest gains on RealX3D. Performance relative to
other methods varies across datasets and metrics, particularly on Phototourism
and in sparse-view settings. Qualitative comparisons are shown in
Fig.~\ref{fig:qualitative_grid_realx3d} for challenging scenes containing
transient occluders and appearance variation. In the examples shown,
WildAnySplat generally reduces transient ghosting and cross-view color
inconsistency relative to the compared methods.



\paragraph{Robustness to Real-World Variation.} WildAnySplat achieves its largest quantitative gains on the RealX3D dataset. At 9 views, it achieves a PSNR of 19.83\,dB, approximately 2\,dB above the
next-best method, WildSplatter, while obtaining an LPIPS of 0.270. 


\paragraph{Backbone Improvement.} Compared with its AnySplat backbone, WildAnySplat improves most evaluated metrics and settings. For instance, on NeRF On-the-Go with 12 input views, WildAnySplat improves
PSNR from 14.14\,dB to 15.24\,dB relative to AnySplat. The component ablation
in the supplementary material further examines the individual contributions of
transient-aware fusion and bilateral-grid appearance modeling. 

\paragraph{Effect of Input-View Count.}
The relative ranking varies with the number of input views. Off-the-Grid is competitive in sparse-view settings, such as the 3-view
RobustNeRF evaluation. WildAnySplat benefits from additional input views and
becomes more competitive on RobustNeRF and NeRF On-the-Go, consistent with the use of cross-view evidence for transient
suppression.


\paragraph{Challenging Illumination.} On the Phototourism benchmark, which is characterized by extreme appearance changes, WildAnySplat remains competitive. Although WildSplatter performs best in SSIM and LPIPS at higher view counts, WildAnySplat achieves comparable PSNR (e.g. 12.57 dB at 3 views).

\paragraph{Visual and Quantitative Discrepancy.}
Although WildAnySplat often reduces transient ghosting and
cross-view color inconsistency, these improvements are not
always reflected by full-image metrics. Transient artifacts often occupy only a small fraction of the image, thereby having limited effect on image-averaged metrics. Moreover, these metrics remain sensitive to small pose
and alignment errors, which are common for sparse, unposed in-the-wild inputs. We therefore interpret the quantitative results together with the qualitative comparisons and do not claim uniform improvement over every baseline or evaluation setting.

\section{Limitations}

Despite strong performance, WildAnySplat has several limitations. First, although our model improves robustness to distractors, it can still struggle in scenes containing large or persistent dynamic regions. When moving objects dominate the scene, pose and reconstruction quality can degrade due to insufficient static observations. 

Second, our bilateral-grid appearance model primarily captures low-frequency spatially-varying color transformations. Extreme illumination changes, strong view-dependent effects, or highly non-linear camera processing may not be fully modeled, leading to residual appearance inconsistencies.

Third, we rely on optimization-based methods to generate teacher supervision for real-world scenes. Although this supervision is computed offline, generating pseudo-clean ground truth introduces additional preprocessing cost and the student may inherit biases or artifacts from the teacher. 

Finally, performance remains limited on highly challenging scenes with severe
multi-view inconsistencies. Addressing these cases may require stronger
geometric reasoning or explicit dynamic-scene modeling.


\section{Conclusion}

We have presented \textit{WildAnySplat}, a feed-forward framework for robust 3DGS from in-the-wild images. Our method addresses two key challenges in unconstrained multi-view reconstruction: transient occlusions and cross-view appearance variation.

To achieve this, we introduced transient-aware masking for robust Gaussian aggregation and supervision, bilateral-grid appearance modeling for canonical color reconstruction, and a student--teacher training framework that leverages clean supervisory signals derived from synthetic augmentation and optimization-based methods.

Experiments on challenging in-the-wild datasets show that the proposed
components improve the robustness of the AnySplat backbone and provide
competitive performance among feed-forward 3DGS methods across varied capture
conditions.

\bibliographystyle{ACM-Reference-Format}
\bibliography{wildanysplat}

\clearpage
\appendix

\begin{figure*}[t]
    \centering
    \includegraphics[width=0.9\textwidth]{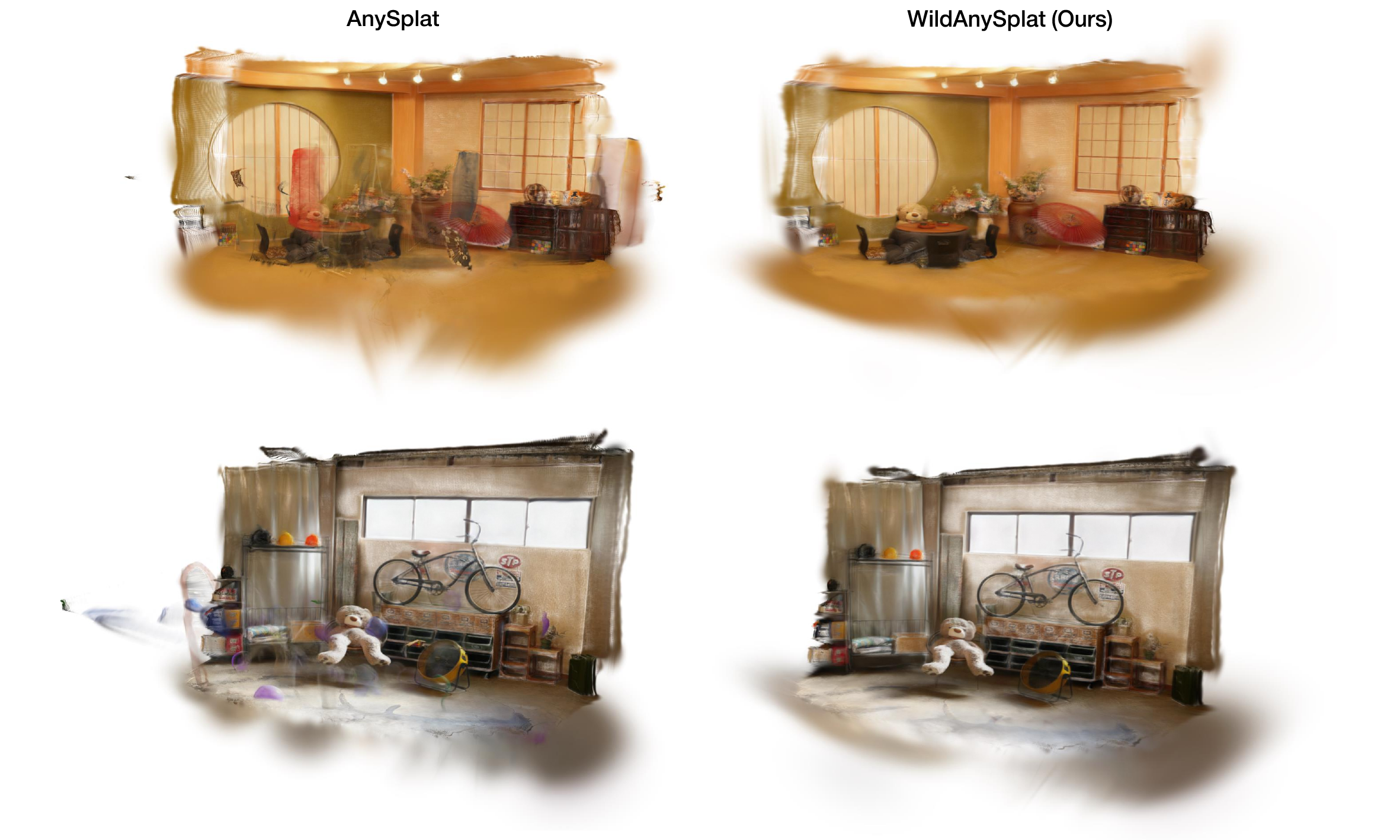}
\caption{
Qualitative 3D reconstruction comparison on scenes from the RealX3D~\cite{realx3d} dataset.}
\label{fig:qualitative_3d_grid_realx3d}
\end{figure*}

\begin{figure*}
    \centering
    \includegraphics[width=0.65\textwidth]{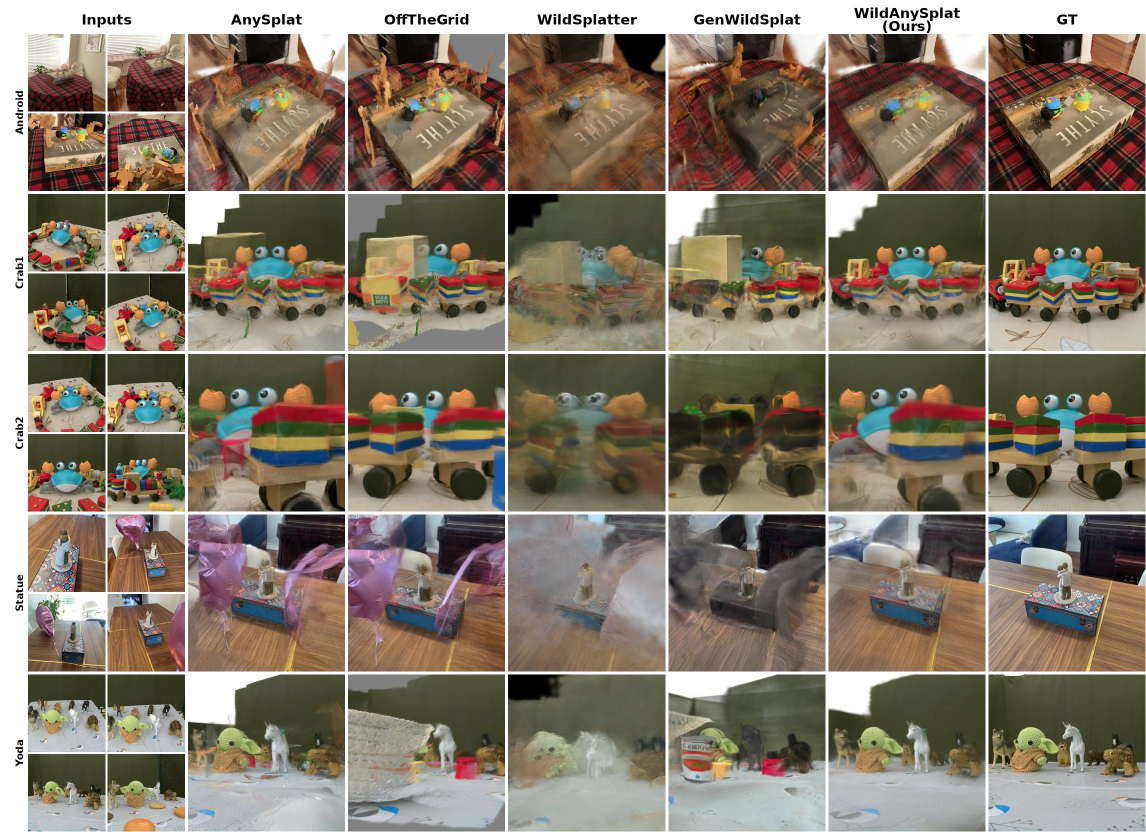}
    \caption{
    Qualitative NVS comparison on scenes from the Robust NeRF~\cite{robustnerf} dataset.
    }
\label{fig:qualitative_grid_robustnerf}
\end{figure*}

\begin{figure*}
    \centering
\includegraphics[width=0.7\textwidth]{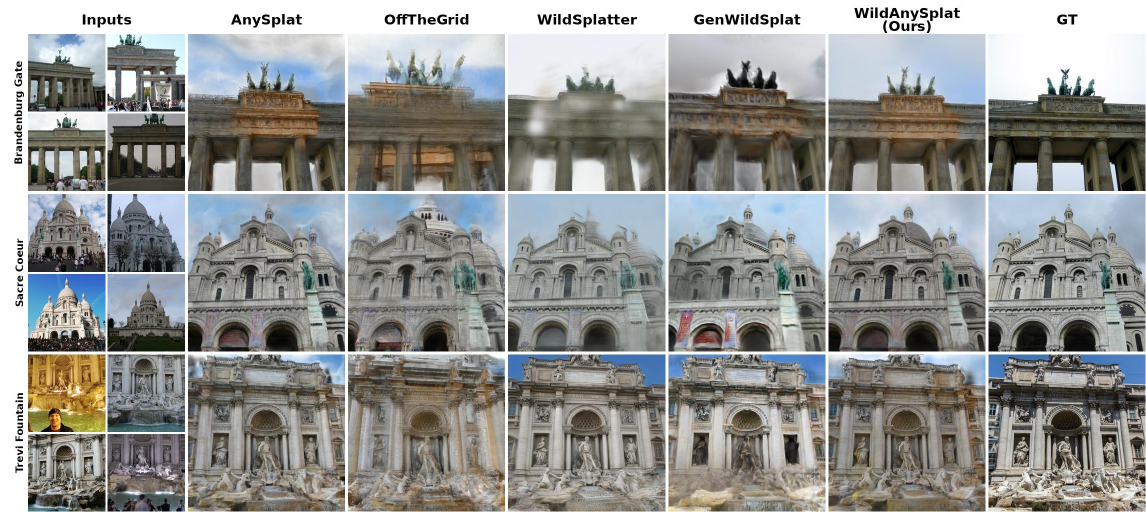}
    \caption{
    Qualitative NVS comparison on scenes from the Phototourism~\cite{phototourism} dataset.
    }
\label{fig:qualitative_grid_phototourism}
\end{figure*}

\begin{figure*}
    \centering
    \includegraphics[width=0.7\textwidth]{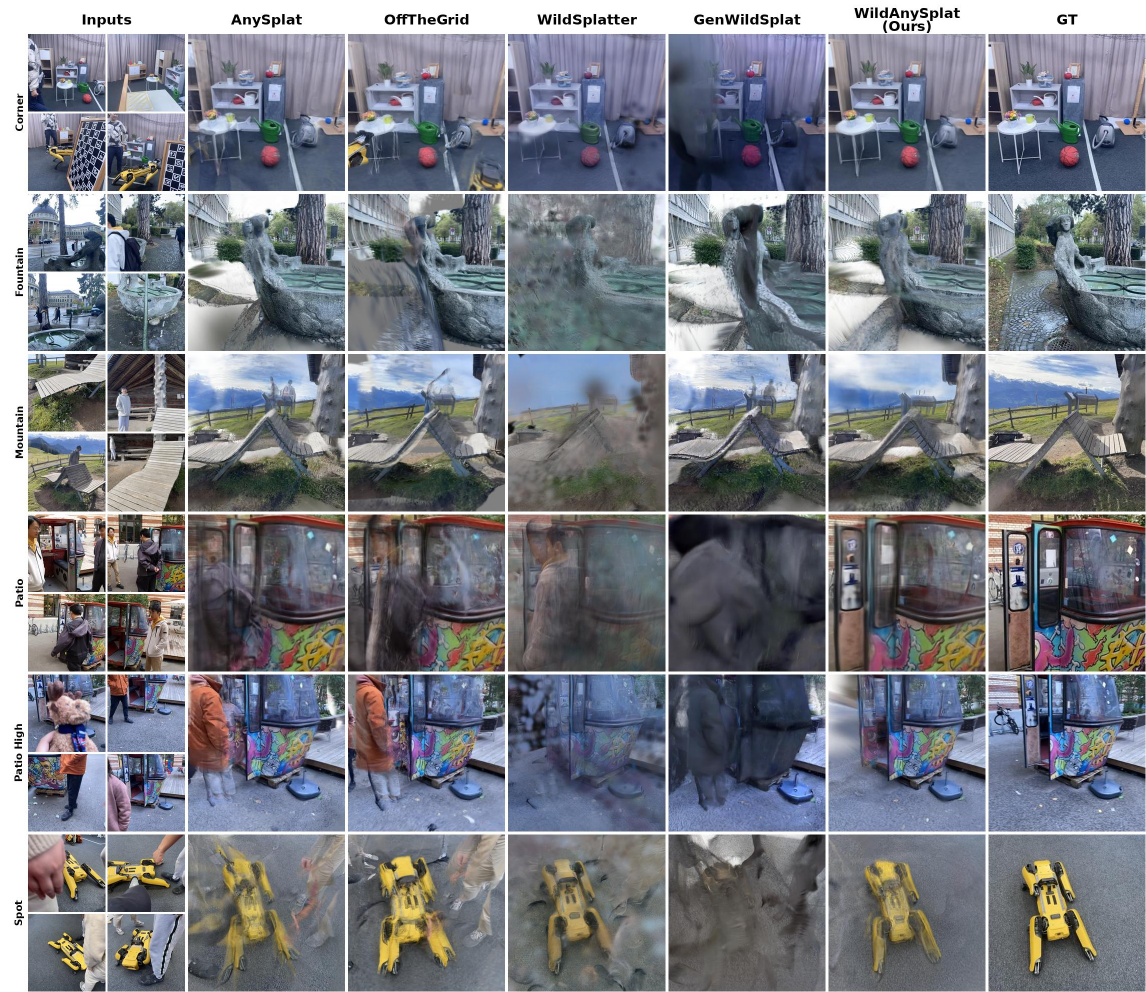}
    \caption{
    Qualitative NVS comparison on scenes from the NeRF On-the-Go~\cite{nerfonthego} dataset.
    }
\label{fig:qualitative_grid_nerfonthego}
\end{figure*}

\clearpage

\end{document}